\pdfoutput=1

\documentclass[11pt]{article}

\usepackage{emnlp2022} 
\usepackage{booktabs}       
\usepackage{amsmath}
\DeclareMathOperator*{\argmin}{argmin}
\usepackage{graphicx}
\usepackage{subcaption}
\usepackage{dblfloatfix}
\usepackage{times}
\usepackage{latexsym}
\usepackage{amsfonts}
\usepackage[T1]{fontenc}

\usepackage[utf8]{inputenc}

\usepackage{microtype}
\newcommand{\minisection}[1]{\noindent{\bf #1}\nobreak}

\title{\textit{N}-Grammer: Augmenting Transformers with latent 
\textit{n}-grams}
\author{%
  {\normalfont Aurko Roy}\thanks{* Equal contributions.}\,\,\,,\,\,
  {\normalfont Rohan Anil}$^*$,\,\, 
  {\normalfont Guangda Lai},\,\,  
  {\normalfont Benjamin Lee},\,\,
  {\normalfont Jeffrey Zhao},\,\,   \\  
  {\normalfont Shuyuan Zhang}\,\,
  {\normalfont Shibo Wang},\,\,  
  Ye Zhang,\,\, 
  Shen Wu,\,\,  
  Rigel Swavely,\,\,  
  Tao (Alex) Yu,\,\,   \\
  Phuong Dao,\,\,  
  Christopher Fifty,\,\, 
  Zhifeng Chen,\,\,  
  Yonghui Wu\,\,  \\
  \\
  {Google Research, Brain Team, Mountain View, CA}\\
  \texttt{\{aurkor, rohananil\}@google.com} \\
}

\begin{document}
\maketitle
\begin{abstract}
Transformer models have recently emerged as one of the foundational models in natural language processing, and as a byproduct, there is significant
recent interest and investment in scaling these models. However, the 
training and inference costs of these large Transformer language models 
are prohibitive, thus
necessitating more research in identifying more efficient variants.
In this work, we propose a simple yet effective modification to the Transformer architecture 
inspired by the literature in statistical language modeling, 
by augmenting the model with \textit{n}-grams that are constructed from a 
discrete latent representation of the text sequence. 
We evaluate our model, the \textit{N}-Grammer on language modeling on the 
C4 data-set as well as text classification on the SuperGLUE data-set,
and find that it outperforms several strong
baselines such as the Transformer and the Primer.
We open-source our model for reproducibility purposes in Jax  \footnote{\url{https://github.com/tensorflow/lingvo/tree/master/lingvo/jax}}.
\end{abstract}

\section{Introduction}
The area of generative modeling of text has witnessed rapid and impressive 
progress driven by the adoption of self-attention to neural networks. Attention
for machine translation was proposed in \citet{bahdanau2014neural, cho2014learning, vaswani2017attention} 
and subsequent works such as  \citet{radford2018improving,devlin2018bert} applied the learned 
representations of language to several problems in natural language processing. The rapid progress has been made possible primarily by increasing the modeling capacity of these Transformer based models to billions of 
parameters  \citep{brown2020language} which comes at a large computational cost. The computational cost of 
Transformer models is being addressed in the literature by exploiting 
sparsity in self-attention  
\citep{ainslie2020etc, zaheer2020big, roy2021efficient},  
mixtures of experts \citep{shazeer2017outrageously,lepikhin2020gshard, fedus2021switch} for sparsity in the feed-forward
network, sparsity in the softmax computation \citep{correia2019adaptively},
and combining depth-wise convolution with attention \citep{wu2021cvt, so2021primer}. 

Motivated by the growing literature in training more efficient variants of Transformers, as well as the
classical literature on statistical language modeling \citep{koehn2009statistical},
we propose a simple modification to the Transformer architecture termed the \textit{N}-Grammer in this work.
The \textit{N}-Grammer layer improves the efficiency of language models by incorporating latent 
\textit{n}-gram representations into the model during training. Since the \textit{N}-Grammer layer only 
involves sparse operations during training and inference, we find that a Transformer model with the latent \textit{N}-Grammer layer can match the quality of a larger Transformer while being significantly faster at inference. This is due to the fact that on most hardware platforms,
the overhead of adding sparse operations such as an embedding look-up
required by the \textit{N}-Grammer is significantly lower
than that of dense matrix multiplication operations incurred by
scaling up the same Transformer model to have the same quality as the \textit{N}-Grammer.

\begin{figure*}[th!]
\centering
\includegraphics[width=0.8\linewidth]{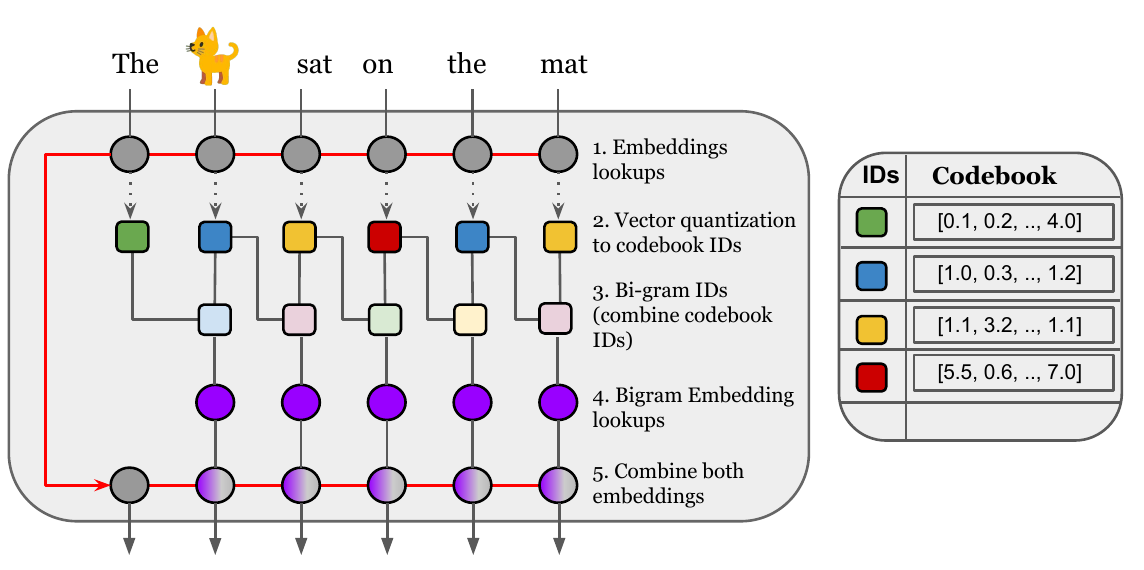}
\caption{The \textit{N}-Grammer layer. It takes as input a 
sequence of uni-gram embeddings and outputs a parallel sequence 
of \textit{N}-gram augmented embeddings. The input embeddings are 
clustered into a discrete latent representation using PQ,
and \textit{n}-grams (bi-grams) IDs are computed over it. 
For each \textit{n}-gram ID, a trainable embedding is looked up 
from an embedding table and combined with 
the input embeddings to produce the output. }\label{fig:ngrammer}
\end{figure*}

\section{Related Work}
\minisection{Memory augmented models}
There has been a long line of work in augmenting sequence models with memory, e.g. the
Neural Turing Machine \citep{graves2014neural} and Memory Networks \citep{weston2014memory}. 
More recent works have proposed 
combining Transformer based models with product key look-up tables \citep{lample2019large},
while \citet{panigrahy2021sketch} propose memories based on sketches of past activations.
There has also been a lot of work on augmenting language models with non-parametric memory,
such as the \(k\)-nearest neighbor language models of \citet{khandelwal2019generalization},
and similar retrieval augmented works such as 
\citet{lewis2020retrieval,guu2020realm,krishna2021hurdles}. 
In these retrieval augmented models, 
the model is conditioned on documents from the training corpus
or a knowledge base, with the hope that information from related articles 
can help improve the factual accuracy of the models.

\minisection{Discrete latent models for sequences}
Discrete latent models using Vector Quantization (VQ) have been widely used in speech
\citep{vqvae, wang2018style,schneider2019wav2vec} 
to learn unsupervised representations of audio signals.
Their use for modeling text sequences were studied in \citet{kaiser2018fast,
roy2018theory} where the motivation was to reduce the inference
latency for neural machine translation models by decoding in the latent space.

\minisection{\textit{N}-gram models for statistical language modeling}
\textit{N}-gram models have a long history in statistical modeling of 
language, see e.g., 
\citet{brown1992class, brown1993mathematics, katz1987estimation,kneser1995improved,chen1999empirical}. 
Before the advent of word vectors and distributed 
representations of language via neural networks
\citep{mikolov2013efficient, wu2016google}, \textit{n}-gram language models were the standard in the 
field of statistical language
modeling. More recent related work on combining neural
sequence models with \textit{n}-gram information is
due to \citet{sun2021revisiting} who propose concatenating
the representations within a local context, while 
\citet{huang2021lookup} propose combining RNN models with
\textit{n}-gram embedding tables. Our work differs from them 
in that we use an \textit{n}-gram look-up table on a 
discrete latent representation of the sequence,
which leads to a more meaningful assignment of shared
\textit{n}-gram representations.

\minisection{Product Quantization}
There has also been a long line of work on investigating variants of Vector
Quantization (VQ) that realize different trade-offs in data compression. 
The most related work in this domain is due to \citet{jegou2011product} who 
introduce a multi-head version of VQ which is termed
Product Quantization (PQ). PQ is widely used in computer 
vision, see e.g., \citet{ge2013optimized, yu2018product}. 
Our approach to learning
discrete latent codes use PQ over the attention heads.

\section{The \textit{N}-Grammer layer}
At a high level, we introduce a simple layer that augments the Transformer architecture with more memory based on latent \textit{n}-grams. While the \textit{N}-Grammer layer is general enough for
considering arbitrary \textit{N}-grams, 
we restrict ourselves to the use of bi-grams. 
We leave the exploration of higher-order \textit{n}-grams for future work.
The layer consists of four core operations: 
\begin{enumerate}
\item Given a sequence of uni-gram embeddings of a text, infer 
a sequence of 
discrete latent representation via PQ. 

\item Infer the bi-gram representation for the latent sequence.

\item Look up trainable bi-gram embeddings via hashing into the bi-gram vocabulary.

\item Combine the bi-gram embeddings with the input uni-gram embeddings.
\end{enumerate}
We describe each of these operations in more detail in the following sections.
To refer to a set of discrete items, we use the notation 
\([m]\) to mean the set \(\{0, 1, \cdots, m-1\}\).

\subsection{Discrete latent representation of a sequence}\label{sec:vq}
The first step of the \textit{N}-Grammer layer is to obtain a parallel sequence of discrete latent
representations with Product Quantization (PQ) \citep{jegou2011product} by learning a codebook from the given sequence of input embeddings. The input embedding is a sequence of uni-gram 
embeddings \(x \in {\mathbb{R}^{l\times h \times d}}\), 
where \(l\) is the length of the sequence, \(h\) is the number of heads, and \(d\) is the embedding dimension per head. We learn a codebook \(c\) in \(\mathbb{R}^{k \times h \times d}\) with \(k\) code-words with 
mini-batch \(k\)-means \citep{bottou1995convergence}, and in the same step, we form the parallel sequence of 
discrete latent representation \(z \in [k]^{l \times h}\) of the sequence \(x\) by picking the codebook IDs 
that have the least distance from the input embeddings:
\[z_{i, j} = \argmin_{l\in [k]} \left\lVert x_{i, j} - c_{l, j} \right\rVert_2.\] 
The advantage of this latent representation \(z\) is twofold. Firstly, it makes considering all \(k^2\) 
bi-grams tractable by mapping the uni-gram embeddings to share the same code-word embedding based on 
similarity, thereby allowing us to use a smaller bi-gram embedding table. 
Secondly, when using a fixed size bi-gram vocabulary, having this latent 
representation allows for a more efficient representation to be learned
compared to directly using the uni-gram IDs. For instance, a uni-gram 
vocabulary of \(32,000\) would entail a bi-gram vocabulary of roughly 
\(1\) billion, which adds a significant memory overhead. 

\subsection{Bi-gram IDs from discrete latent 
representation}\label{sec:compute-bigram}
The second step is to convert the discrete latent 
representation \(z\) 
computed in Section~\ref{sec:vq} to bi-gram IDs \(b \in 
[k^2]^{l\times h}\). The latent 
bi-gram IDs are formed at each
position by combining the uni-gram latent IDs \(z\) from the previous position as
\[b_{i} = \begin{cases}
                z_i & \text{if \(i = 0\),}\\
                z_i + k z_{i-1} & \text{otherwise}
          \end{cases}
\]
where \(k\) is the size of our codebook. 
This directly maps the discrete latent sequence from  a vocabulary space of \([k]\) to the
latent bi-gram vocabulary space of \([k^2]\).

\subsection{Constructing bi-gram representations}\label{sec:bigram} 
The third step is to construct bi-gram latent 
representations \(b\) of
the sequence. We can consider all \(k^2\) bi-grams and 
augment the model with an embedding for each such bi-gram. 
In practice, the compression for machine translation 
models with a  uni-gram vocabulary of \(32,000\) involves 
clustering each token into 
roughly \(k = 2^{12}\) 
clusters without sacrificing quality \citep{kaiser2018fast, roy2018theory}. 
In this instance, to consider all bi-grams would involve 
constructing an embedding table with \(16\) 
million rows. Since this is still large, we map the latent
bi-gram IDs to a smaller bi-gram vocabulary of size \(v\),
by using separate hash functions for each head.

More precisely, we have a latent bi-gram embedding table
\(B \in \mathbb{R}^{v \times h \times d_b}\), 
where \(v\) is the bi-gram vocabulary and \(d_b\) is
the bi-gram embedding dimension. The bi-gram embedding 
\(y \in \mathbb{R}^{l \times h \times d_b}\) of the text sequence
is then constructed as 
\(y_{i, j} = B\left[\left((r_j b_{i, j} + s_j)\mod{p_j}\right) \mod{v}, j\right],\)
where for each head \(j\), we select a random prime \(p_j\) greater than \(k^2\), 
and \(r_j\) is chosen randomly in \(\{1, \cdots, p-1\}\) and
\(s_j\) is chosen randomly in \([p-1]\). 
This scheme is a universal hashing scheme and
guarantees a low collision probability for the discrete
latent codes of each head \citep{thorup2015high}.
Note that the bi-gram embedding vector \(y_{i,j}\) is a \(d_b\)-dimensional vector. 

\begin{table*}[ht]
\centering
\begin{tabular}{lccccccc}
\toprule
       Model &  Layers & Params & Vocab size & Clusters & Dim & Inference Ex/sec & PP \\
       \midrule
        Transformer & 16 & 234M & - & -  & - & 402.00 & 15.32 \\ 
        Transformer-L & 20 & 284M & - & - & - & 331.12 & 14.70 \\\midrule
        Primer & 16 & 234M & - & -  &- & 346.32 & 15.10 \\ 
        Primer-L & 18 & 284M & - & - & - & 284.40  & 15.01 \\ \midrule
        \textit{N}-Grammer & 16 & 246M & 196K & - & 12.5\% & 379.60 &  15.36 \\ 
        \textit{N}-Grammer & 16 & 259M & 196K & - & 25.0\% & 378.64 & 15.27 \\
        \textit{N}-Grammer & 16 & 284M & 196K & - & 50.0\% & 375.40 & 15.50 \\ \midrule
        \textit{N}-Grammer & 16 & 251M & 196K & 4K & 12.5\% & 366.80 & 15.26 \\
        \textit{N}-Grammer & 16 & 263M & 196K & 4K & 25.0\% & 362.40 &  15.07 \\ 
        \textit{N}-Grammer & 16 & 288M & 196K & 4K & 50.0\% & 362.00 & 15.01 \\  \midrule
        \textit{N}-Grammer & 16 & 255M & 196K & 8K & 12.5\% & 359.52 & 15.58 \\ 
        \textit{N}-Grammer & 16 & 267M & 196K & 8K & 25.0\% & 358.96 & 15.44 \\
        \textit{N}-Grammer & 16 & 292M & 196K & 8K & 50.0\% & 358.16 & 15.01 \\ \midrule
        \textit{N}-Grammer & 16 & 267M & 393K & 8K & 12.5\% & 363.60 & 15.32 \\
        \textit{N}-Grammer & 16 & 292M & 393K & 8K & 25.0\% & 360.16 & 15.97 \\ 
        \textit{N}-Grammer & 16 & 343M & 393K & 8K & 50.0\% & 356.94 & 14.79 \\ 
    \bottomrule
    \end{tabular}
\vspace{1mm}
\caption{Ablation results on auto-regressive language modeling on the C4 
data-set \citep{raffel2019exploring}. 
The column labeled \textit{Vocab Size} refers to the bi-gram vocabulary size, 
while the column labeled \textit{Dim} refers to the 
bi-gram embedding dimension as a percentage of the total 
model dimension. All models are trained with a batch size 
of \(256\) for a total of \(1\)M steps. We report the test
perplexity (\textit{PP}) and as well as the inference
through-put in examples per second 
(\textit{Inference Ex/sec}) on a 
TPU-v3 with 8 cores (higher is better).}
\label{tab:c4-ablation}
\end{table*}

\subsection{Combining the embeddings}\label{sec:combining}
The final step is to form a new representation of the text sequence which is derived by combining the uni-gram embedding \(x \in 
\mathbb{R}^{l \times h \times d}\)
with the latent bi-gram embedding \(y \in 
\mathbb{R}^{l\times h \times d_b}\)
obtained in Section~\ref{sec:bigram}. The bi-gram embedding and uni-gram embedding are both independently layer normalized (\(LN\)), followed by simply concatenating the two along the embedding dimension to produce
\(w = [LN(x), LN(y)] \in \mathbb{R}^{l \times h \times (d + d_b)}\) which is passed as input to rest of the Transformer network. 
Note that layer normalization \citep{ba2016layer} leads to more stable training. 

\section{Experiments \& Results}
We compare the \textit{N}-Grammer model with the Transformer architecture \citep{vaswani2017attention} 
as well as with the recently proposed Primer architecture \citep{so2021primer} on the C4 data-set
\citep{raffel2019exploring}
\footnote{\url{https://www.tensorflow.org/datasets/catalog/c4}}. To establish a strong baseline for our experiments we use a Gated Linear Unit
\citep{dauphin2017language} as the feed-forward network with a GELU activation function 
\citep{hendrycks2016gaussian} in all our models, except the Primer. The Primer architecture 
uses a \(3\times 1\) depth-wise convolution after the key,
query and value projections, and the squared RELU 
activation function as proposed in 
\citet{so2021primer}. For all experiments, we use the rotary position embedding (RoPE) from \citet{su2021roformer},
which greatly improves the quality of all models. 

We compare the \textit{N}-Grammer, Primer and Transformer models in Table~\ref{tab:c4-ablation}. The baseline Transformer model has \(16\) layers and \(8\) heads, with a model dimension of \(1024\). We train all the models
with a batch size of \(256\) and a sequence length of \(1024\) on a TPU-v3. A more detailed exposition of the various
hyper-parameter choices is given in Section~\ref{sec:hparams}.
For the \textit{N}-Grammer models, we ablate 
with different sizes for the bi-gram embedding dimension ranging from \(128\)
to \(512\). Since adding \textit{n}-gram embeddings increases 
the number of trainable parameters, we also train two large 
baselines in Table~\ref{tab:c4-ablation} (Transformer-L  and 
Primer-L) which have the same order of parameters as the 
\textit{N}-Grammer models. However, unlike the larger 
Transformer models, the training and inference cost of 
\textit{N}-Grammer does not scale 
proportional to the number of parameters in the
\textit{n}-gram embedding layer, 
since they rely on sparse look-up operations 
(see column Inference Ex/sec in 
Table~\ref{tab:c4-ablation}). Thus for example, we find from 
Table~\ref{tab:c4-ablation} that the best \textit{N}-Grammer
model with a \textit{n}-gram vocabulary of \(393\)K and a discrete
latent vocabulary of \(8\)K matches the quality of Transformer-L
and Primer-L in perplexity (14.79 vs 14.70 vs 15.01) while having
significantly higher through-put (356.94 vs 331.12 vs 284.40 
examples/sec).

We also examine a simple version of \textit{N}-Grammer where we compute the \textit{n}-grams
directly from the uni-gram vocabulary as in Section~\ref{sec:bigram} rather than from the latent 
representation of Section~\ref{sec:vq}. This is reported in 
Table~\ref{tab:c4-ablation} and corresponds to the 
\textit{N}-Grammer without an entry
in the clusters column. Note that in this case, the modulo 
hashing scheme of 
Section~\ref{sec:bigram} is random and independent of the 
content of the actual uni-gram embeddings. We inspect the 
individual cluster assignment in Section~\ref{sec:latent_rep} 
and find common themes among the groupings.

\section{Hyper-parameters for experiments}\label{sec:hparams}
In this section we report the hyper-parameter settings for all our experiments for reproducibility purposes.

\subsection{Optimizer hyper-parameters}
We use the Adam optimizer
\citep{adam} and tune the learning rate as well as
\(\varepsilon\) as reported in 
\citep{agarwal2020disentangling}. 
We find that decreasing \(\varepsilon\) from the standard 
setting of \(10^{-6}\) to 
\(10^{-10}\) benefits the Transformer models while having less of an effect on the
Primer \citep{so2021primer}. We use a learning rate of 
\(10^{-3}\) for all models.
We use a \(\beta_1 = 0.9\) and \(\beta_2 = 0.99\) and clip the gradient norm to \(5.0\).
We do not use any weight decay. We train all models with a global batch size of \(256\) on a TPU-v3 with \(32\)
cores and a sequence
length of \(1024\).

\subsection{\textit{N}-Grammer hyper-parameters}
For the \textit{N}-Grammer models, we use a discrete latent vocabulary of \(k=\{4096,8192\}\)
except for the baseline \textit{N}-Grammer models which directly compute \textit{n}-grams
on the uni-gram vocabulary. For training the 
\textit{n}-gram embedding tables we use the Adagrad 
optimizer~\citep{duchi2011adaptive}, which is known to be
more suitable for learning sparse features. We use a learning 
rate of \(0.1\) for training the \textit{n}-gram embedding table,
with the same learning rate schedule as the base model. We find that
using a \(10\times\) higher or lower learning rate leads to unstable
training of the \textit{N}-Grammer model.

We train the cluster centers for learning the discrete latent 
representation using mini-batch \(k\)-means 
\citep{bottou1995convergence}. We do not use
any smoothing or exponential moving averages for either the counts or the centers, since we find 
empirically that it doesn't help in our setting. 
We use a learning rate of \(10^{-3}\) for learning the
discrete representations.

\section{Position of the \textit{N}-Grammer layer}\label{sec:position_ablations}
We perform ablation experiments on the position of the
latent \textit{N}-Grammer layer, 
since potentially one may add it to any intermediate layer
of the network. We take the best \textit{N}-Grammer
model from Table~\ref{tab:c4-ablation}, corresponding to
an \textit{n}-gram vocabulary size of \(393\)K and a latent
vocabulary of size \(8\)K and ablate the position of
the \textit{N}-Grammer layer in Table~\ref{tab:position-ablation}. We observe that placing
the \textit{n}-gram layer at the beginning of the network
turns out to be the best choice, since moving the layer 
successively to the end of the network leads to progressively worse performance. We hypothesize
that this is due to the presence of 
fewer attention layers to leverage the improved 
representations from the \(n\)-gram embeddings.
\begin{table}[ht]
\centering
\begin{tabular}{lc}
\toprule
       Model &  PP \\
       \midrule
        \textit{N}-Grammer & 14.79 \\ 
        \textit{N}-Grammer \textit{begin} & 14.92 \\ 
        \textit{N}-Grammer \textit{mid} & 15.13\\
        \textit{N}-Grammer \textit{end} & 15.17\\
    \bottomrule
    \end{tabular}
    \vspace{1mm}
    \caption{Ablation results on the position of the 
    \textit{N}-Grammer layer on the C4 
    data-set~\citep{raffel2019exploring}. We use the best
    \textit{N}-Grammer model from Table~\ref{tab:c4-ablation},
    with a vocab size of \(393\)K. The \textit{N}-Grammer
    models labelled \textit{begin}, \textit{middle} and \textit{end} refer to the latent \textit{n}-gram 
    embedding layer being placed after the first layer,
    the middle layer and the end layer respectively.}
    \label{tab:position-ablation}
\end{table}

\section{Optimizing through-put}
We note that there is a trade-off in computing the discrete 
latent representation of a text sequence, 
where it may be more efficient in practice to cluster the 
uni-gram vocabulary directly rather than clustering the 
embedded text sequence. This is an important consideration 
when serving the \textit{N}-Grammer model, since the mapping 
from token to discrete latent is fixed after the completion
of training thereby allowing us to pay a one time cost
in computing this mapping for the entire vocabulary. 
We formulate this more precisely as follows.

Let the uni-gram vocabulary be \(v\), the latent vocabulary
\(k\), the sequence length \(l\), batch size \(b\),
and let the  \textit{N}-Grammer model serve a 
total of \(m\) examples. 
If we were to compute the latent representation for each
sequence, we incur a cost of \(\mathcal{O}(bkl)\) per sequence. 
On the other hand, if we were to compute
the latent representation for the entire vocabulary up-front
and cache the mapping from token to latent, we pay a one time 
cost of \(\mathcal{O}(vk)\) for inferring the latent representations. 
Assuming an \(\mathcal{O}(1)\) cost of looking up the latent 
representation per sequence, this
cost can be amortized over the \(m\) examples to get a per 
sequence cost of \(\mathcal{O}(vk/m + 1)\). As \(m \rightarrow \infty\),
i.e., the model is continuously deployed, 
we essentially get to compute the discrete latent 
representation in constant time per sequence during serving.

Since computing the \textit{n}-gram ID (see 
Section~\ref{sec:compute-bigram})
and retrieving the \textit{n}-gram
representations are also constant (with respect to the
number of attention layers, attention 
heads and model dimension)
time operations per sequence, this implies that when
served long enough, 
the \textit{N}-Grammer model essentially incurs a 
constant overhead over the Transformer.

\section{Convergence comparisons}
\begin{figure*}[htbp]
\centering
    \begin{subfigure}[l]{1.0\columnwidth}
        \includegraphics[width=1.0\textwidth]{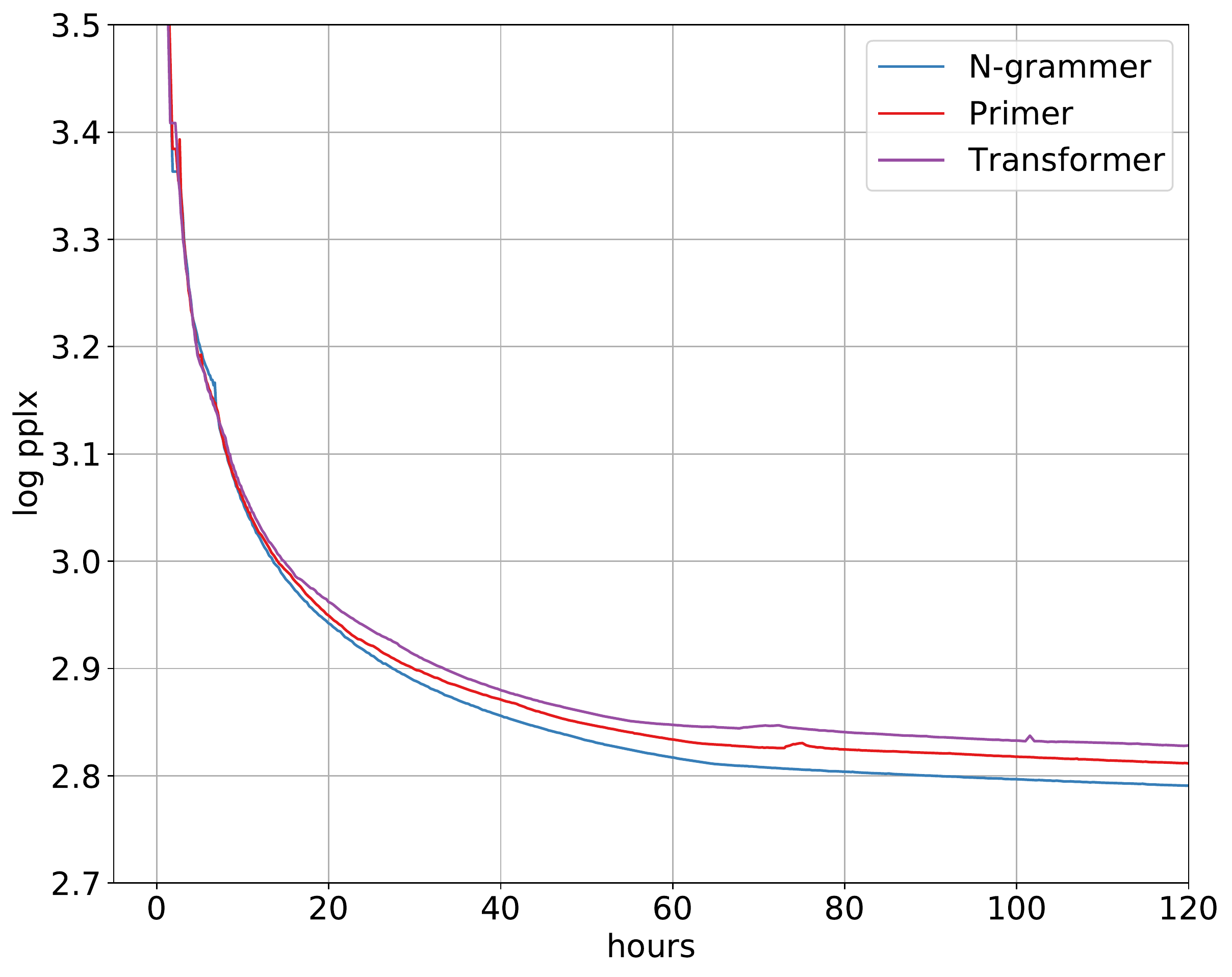}
        \caption{Log perplexity vs wall-clock time on TPU-v3}\label{fig:wall-ppl}
    \end{subfigure}
    \hfill{}
    \begin{subfigure}[r]{1.0\columnwidth}
        \includegraphics[width=1.0\textwidth]{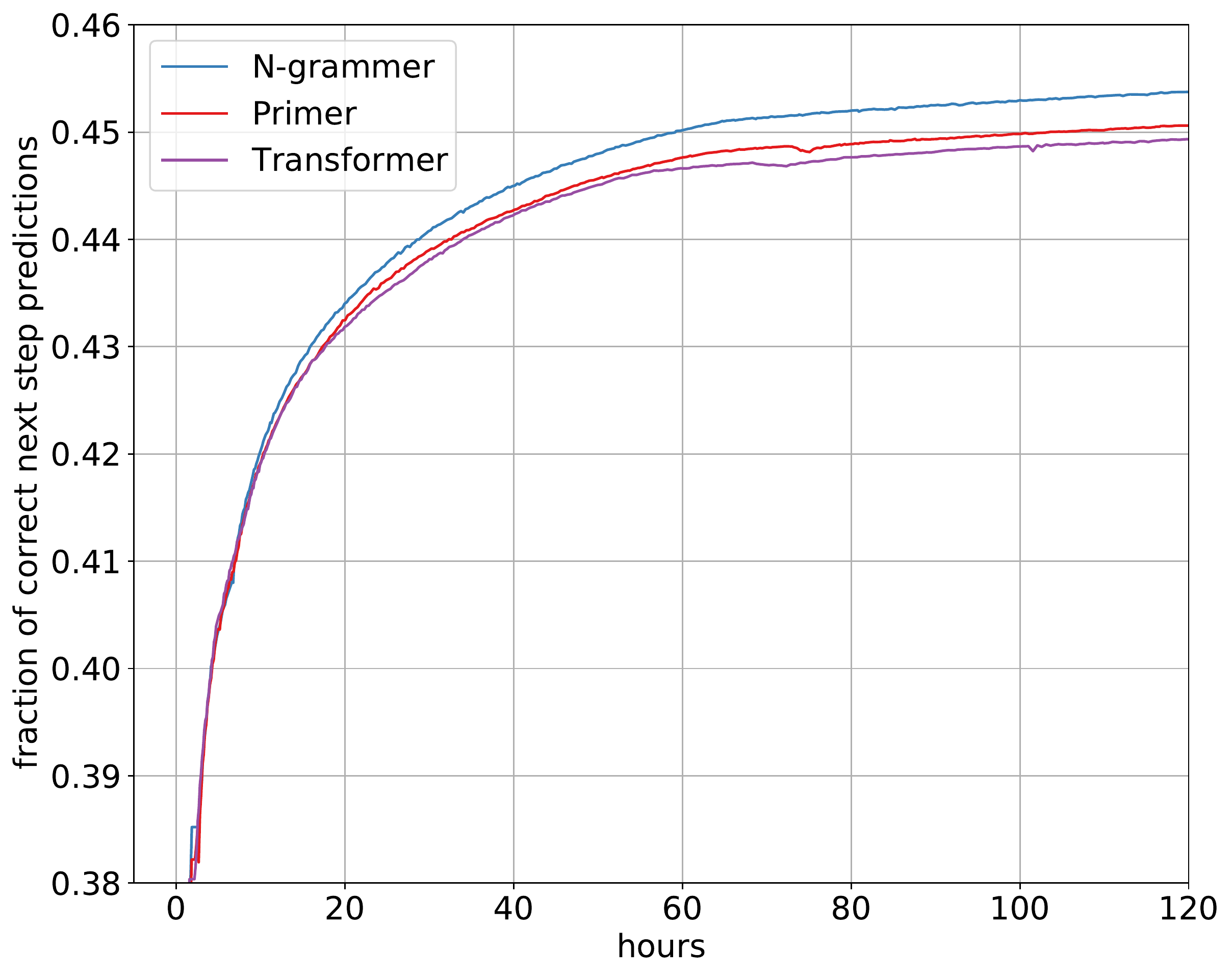}
        \caption{Top-1 accuracy vs wall-clock time on TPU-v3}\label{fig:wall-acc}
    \end{subfigure}
\caption{Wall-clock time comparisons between Transformer with 
Gated GELU, Primer and \textit{N}-Grammer on the C4 data-set \citep{raffel2019exploring}.}
\label{fig:wall}
\end{figure*}

We have included training curve comparisons of the \textit{N}-Grammer with that of
the Transformer \citep{vaswani2017attention} and the Primer \citep{so2021primer}. We compare the
three models in Figures~\ref{fig:wall-ppl} 
and~\ref{fig:wall-acc} where the \(x\)-axis denotes the wall clock
time on a TPU-v3 while the \(y\)-axis denotes the log perplexity and top-1 accuracy respectively
on the C4 data-set \citep{raffel2019exploring}. 
From Figure~\ref{fig:wall} we see that the 
\textit{N}-Grammer model is roughly \(2\times\) faster than the Primer in wall clock time to reach
the same perplexity or accuracy. More precisely, the
baseline Primer model after 1M steps (180 TPU hours)
has a perplexity of
\(15.10\), which the 
best \textit{N}-Grammer model from Table~\ref{tab:c4-ablation}
achieves at \(465\)K steps (90 TPU hours).

\section{Comparison on downstream tasks}\label{sec:super-glue}
We fine-tune the baseline Transformer, Primer (as well as their 
large variants) and the best \textit{N}-Grammer model from 
Table~\ref{tab:c4-ablation}
on the SuperGLUE benchmark~\citep{wang2019superglue} to evaluate
whether the perplexity gains of the \textit{N}-Grammer model
also result in downstream classification gains.
For all models we take the checkpoint at \(1\)M steps,
and fine-tune for \(100\)K steps with a constant learning
rate of \(10^{-4}\). We report the downstream evaluation
metrics in Table~\ref{tab:superglue}. 
From Table~\ref{tab:superglue} we observe that the \textit{N}-Grammer
improves on the quality of the Transformer and Primer models on most 
SuperGLUE tasks. More surprisingly, we see that it also substantially
improves on the larger Transformer-L and Primer-L models on tasks like
COPA, RTE, WiC and WSC.
\begin{table*}[htbp]
\centering
\resizebox{\textwidth}{!}{
\begin{tabular}{lccccccccc}
\toprule
       Model & BoolQ & CB & COPA & MultiRC & ReCoRD
       & RTE & WiC & WSC & Avg\\
       & Acc. & F1/Acc. & Acc. & F1/EM & F1/EM & Acc. & Acc.
       & Acc. & \\
       \midrule
        Transformer & 65.23 & 54.03/67.86 & 55.0 & 59.47/13.64  & 30.34/29.32 & 53.43 & 54.55 & 61.54 & 52.13\\
        Transformer-L & 66.15 & \textbf{73.74}/\textbf{75.00} & 59.0
        & 62.09/12.17 & 29.19/28.21 & 57.04 & 55.33 & 62.50 & \textbf{55.03}\\
        \midrule
        Primer & \textbf{66.27} & 62.64/71.43 & 53.0 & \textbf{62.89}/\textbf{12.91} 
        & \textbf{30.56}/\textbf{29.55} & 55.96 & 54.86 & 65.38 & 53.81\\
        Primer-L & 62.20 & 47.69/58.93 & 58.0 & 51.83/4.62 & 
        25.21/24.35 & 51.99 & 54.08 & 63.46 & 49.50\\
        \midrule
        \textit{N}-Grammer & 64.98 & 59.69/67.86 & \textbf{60.0} & 61.95/11.33 & 29.90/28.91 & \textbf{59.21} & \textbf{56.11} & \textbf{68.27} & 54.80\\ 
    \bottomrule
    \end{tabular}}
    \caption{Fine-tuning results on 
    SuperGLUE~\citep{wang2019superglue} comparing the 
    Transformer, Primer and the best \textit{N}-Grammer
    model from Table~\ref{tab:c4-ablation}. The 
    \textit{N}-Grammer model has a discrete latent vocabulary
    of size \(8\)K and a \textit{n}-gram vocabulary of
    size \(393\)K. For all models we take the pre-trained 
    checkpoint at 1M steps and 
    fine-tune for \(100\)K steps with a constant
    learning rate of \({10}^{-4}\).}
    \label{tab:superglue}
\end{table*}

\section{Analysis of the latent representations}\label{sec:latent_rep}
We inspect the discrete latent representations learned by the \textit{N}-Grammer layer by
examining the different uni-gram tokens that are assigned to the same cluster ID. 
We take a trained \textit{N}-Grammer model with \(8192\) clusters, \textit{n}-gram embedding
dimension of \(16\) and \textit{n}-gram vocabulary of 196K.
We pass the entire set of \(32,000\) uni-gram embeddings as input to the \textit{N}-Grammer layer, 
thereby gathering the cluster assignment of every uni-gram token.
We present some of these in Table~\ref{tab:cluster},
where we find that the model learns to group related uni-gram tokens together:
\begin{enumerate}
\item the cluster with head ID \(0\) and cluster ID \(6259\) corresponds to sports and games,

\item the cluster with head ID \(2\) and cluster ID \(5362\) corresponds to places,

\item the cluster with head ID \(0\) and cluster ID \(7468\) corresponds to animals and fruits,

\item the cluster with head ID \(2\) and cluster ID \(8080\) corresponds to the arts,

\item the cluster with head ID \(4\) and cluster ID \(6618\) also corresponds to the arts.
\end{enumerate}
We also observe that several heads independently learn a similar themed grouping, e.g., head \(2\) and
\(4\) both have a cluster dedicated to arts and entertainment.
\setlength{\arrayrulewidth}{0.5mm}
\setlength{\tabcolsep}{18pt}
\renewcommand{\arraystretch}{1.5}
\begin{table*}[htbp]
\centering
\begin{tabular}{|p{0.1\linewidth}|p{0.1\linewidth}|p{0.5\linewidth}|}
\hline
       Head ID & Cluster ID & Uni-gram Tokens \\
       \hline
        0 & 6259 & Baseball, football, ceramic, Galaxy, hockey, basketball, Cricket, Basketball, guitar, acquisition, athlete, Soccer, Squid, sports\\\hline
        2 & 5362 & Alchemist, Vegas, hanger, Seinfeld, Kenya, Heroic, Kurdish, Rodgers, Bolivia, Venom, Qatar, dosage, Arcade, Emperor, becua, Finnish, Taiwanese, Chennai, hood, dub, flake, Balkan, Psalm, Bueno, Moldova, flow, mosquito, Filipino, Throne, Siberia, Trout, Fist, Czech, Boulevard, Azerbaijan, Peru, OW, plaster, Kashmir, NZ, Priest, Palestinian, Tibetan, stencil, Aragon, coils, HBO, Iceland, strains, Zimbabwe, firewall, Nepal, Elves, Iranian, Mongol, Traffic, Camilla, parade, Afghan, hose, Serpent, Tarantino, web, Khal, Squid, Mala, Syrian, hood\\\hline
        0 & 7468 & Unknown, spoon, Shut, coconut, grapefruit, cran, Kami, moon, spider, yogurt, perfume, Wine, Skate, antique, snail, Onion, guinea, puppy, mineral, Reagan, elbow, bark, patio, beneath, snake, lever, bunny, falcon, rail, ribbon, knob, apples, quarry, corn, nach, hiking, invoice, Pour, flora, fishing, Paint, olive, violin, octopus, horizontal, blanket, circular, army, nickel, cattle, potato, dolphin, mosquito, citrus, shutter\\\hline
        2 & 8080 & Knicks, Shakespeare, SPE, nursing, spells, Alexa, arrow, vocalist, rehearsal, tunnel, eine, Critical, clar, BAN, remix, obstacle, musicians, BRO, legislature, EMS, Manga, piano, sword, vocal, bald, choir, Messi, Beta, cad, illustrator, organ, conjunction, lunar, bien, needles, musician, hiking, tad, poe, Pay, violin, Marxist, literary, Theater, gig, poetry, Illustrator, guitar, Pluto, Camaro, Fog, orbit, dancing, epub \\\hline
        4 & 6618 & Wise, vocalist, actor, cheek, musicians, TION, piano, tunes, choir, filmmaker, musician, Suzuki, violin, Theater, gig, Drama, guitar, logic, Entertainment\\
\hline
\end{tabular}
\vspace{1mm}
\caption{Mapping of uni-gram tokens to cluster IDs for the \textit{N}-Grammer
model. The \textit{N}-Grammer model has \(8\) heads, \(8192\) clusters, an \textit{n}-gram embedding 
dimension of 16 and a \textit{n}-gram vocabulary of 196K. We report the head index (Head ID), the cluster
index (Cluster ID) and the uni-gram tokens assigned to those IDs for a random subset of clusters.}
\label{tab:cluster}
\end{table*}


\section{Conclusion}
We introduced the \textit{N}-Grammer layer for augmenting 
the Transformer architecture with latent \textit{n}-grams, 
and find that it can match a larger Transformer and Primer
in quality while being significantly faster in 
inference. The \textit{N}-Grammer architecture is particularly
suitable for devices that allow storing large embedding tables
while supporting only distributed gather-scatter operations. 
We also showed that by caching the mapping from token to discrete latent, 
one can serve the \textit{N}-Grammer architecture with only a constant 
overhead over the Transformer. This makes the \textit{N}-Grammer
attractive for deployment, since on most hardware platforms
sparse operations such as an embedding look-up is significantly faster
than dense operations such as matrix multiplications.

\clearpage
\bibliography{anthology,custom}

\begin{thebibliography}{52}
\expandafter\ifx\csname natexlab\endcsname\relax\def\natexlab#1{#1}\fi

\bibitem[{Agarwal et~al.(2020)Agarwal, Anil, Hazan, Koren, and
  Zhang}]{agarwal2020disentangling}
Naman Agarwal, Rohan Anil, Elad Hazan, Tomer Koren, and Cyril Zhang. 2020.
\newblock Disentangling adaptive gradient methods from learning rates.
\newblock \emph{arXiv preprint arXiv:2002.11803}.

\bibitem[{Ainslie et~al.(2020)Ainslie, Ontanon, Alberti, Cvicek, Fisher, Pham,
  Ravula, Sanghai, Wang, and Yang}]{ainslie2020etc}
Joshua Ainslie, Santiago Ontanon, Chris Alberti, Vaclav Cvicek, Zachary Fisher,
  Philip Pham, Anirudh Ravula, Sumit Sanghai, Qifan Wang, and Li~Yang. 2020.
\newblock Etc: Encoding long and structured inputs in transformers.
\newblock \emph{arXiv preprint arXiv:2004.08483}.

\bibitem[{Ba et~al.(2016)Ba, Kiros, and Hinton}]{ba2016layer}
Jimmy~Lei Ba, Jamie~Ryan Kiros, and Geoffrey~E Hinton. 2016.
\newblock Layer normalization.
\newblock \emph{arXiv preprint arXiv:1607.06450}.

\bibitem[{Bahdanau et~al.(2015)Bahdanau, Cho, and Bengio}]{bahdanau2014neural}
Dzmitry Bahdanau, Kyunghyun Cho, and Yoshua Bengio. 2015.
\newblock Neural machine translation by jointly learning to align and
  translate.
\newblock In \emph{3rd International Conference on Learning Representations,
  ICLR 2015}.

\bibitem[{Bottou and Bengio(1995)}]{bottou1995convergence}
Leon Bottou and Yoshua Bengio. 1995.
\newblock Convergence properties of the k-means algorithms.
\newblock In \emph{Advances in neural information processing systems}, pages
  585--592.

\bibitem[{Brown et~al.(1993)Brown, Della~Pietra, Della~Pietra, and
  Mercer}]{brown1993mathematics}
Peter~F Brown, Stephen~A Della~Pietra, Vincent~J Della~Pietra, and Robert~L
  Mercer. 1993.
\newblock The mathematics of statistical machine translation: Parameter
  estimation.
\newblock \emph{Computational linguistics}, 19(2):263--311.

\bibitem[{Brown et~al.(1992)Brown, Della~Pietra, Desouza, Lai, and
  Mercer}]{brown1992class}
Peter~F Brown, Vincent~J Della~Pietra, Peter~V Desouza, Jennifer~C Lai, and
  Robert~L Mercer. 1992.
\newblock Class-based n-gram models of natural language.
\newblock \emph{Computational linguistics}, 18(4):467--480.

\bibitem[{Brown et~al.(2020)Brown, Mann, Ryder, Subbiah, Kaplan, Dhariwal,
  Neelakantan, Shyam, Sastry, Askell et~al.}]{brown2020language}
Tom~B Brown, Benjamin Mann, Nick Ryder, Melanie Subbiah, Jared Kaplan, Prafulla
  Dhariwal, Arvind Neelakantan, Pranav Shyam, Girish Sastry, Amanda Askell,
  et~al. 2020.
\newblock Language models are few-shot learners.
\newblock \emph{arXiv preprint arXiv:2005.14165}.

\bibitem[{Chen and Goodman(1999)}]{chen1999empirical}
Stanley~F Chen and Joshua Goodman. 1999.
\newblock An empirical study of smoothing techniques for language modeling.
\newblock \emph{Computer Speech \& Language}, 13(4):359--394.

\bibitem[{Cho et~al.(2014)Cho, van Merri{\"e}nboer, Gulcehre, Bahdanau,
  Bougares, Schwenk, and Bengio}]{cho2014learning}
Kyunghyun Cho, Bart van Merri{\"e}nboer, Caglar Gulcehre, Dzmitry Bahdanau,
  Fethi Bougares, Holger Schwenk, and Yoshua Bengio. 2014.
\newblock Learning phrase representations using rnn encoder--decoder for
  statistical machine translation.
\newblock In \emph{Proceedings of the 2014 Conference on Empirical Methods in
  Natural Language Processing (EMNLP)}, pages 1724--1734.

\bibitem[{Correia et~al.(2019)Correia, Niculae, and
  Martins}]{correia2019adaptively}
Gon{\c{c}}alo~M Correia, Vlad Niculae, and Andr{\'e}~FT Martins. 2019.
\newblock Adaptively sparse transformers.
\newblock \emph{arXiv preprint arXiv:1909.00015}.

\bibitem[{Dauphin et~al.(2017)Dauphin, Fan, Auli, and
  Grangier}]{dauphin2017language}
Yann~N Dauphin, Angela Fan, Michael Auli, and David Grangier. 2017.
\newblock Language modeling with gated convolutional networks.
\newblock In \emph{International conference on machine learning}, pages
  933--941. PMLR.

\bibitem[{Devlin et~al.(2019)Devlin, Chang, Lee, and
  Toutanova}]{devlin2018bert}
Jacob Devlin, Ming-Wei Chang, Kenton Lee, and Kristina Toutanova. 2019.
\newblock Bert: Pre-training of deep bidirectional transformers for language
  understanding.
\newblock In \emph{NAACL-HLT (1)}.

\bibitem[{Duchi et~al.(2011)Duchi, Hazan, and Singer}]{duchi2011adaptive}
John Duchi, Elad Hazan, and Yoram Singer. 2011.
\newblock Adaptive subgradient methods for online learning and stochastic
  optimization.
\newblock \emph{Journal of machine learning research}, 12(7).

\bibitem[{Fedus et~al.(2021)Fedus, Zoph, and Shazeer}]{fedus2021switch}
William Fedus, Barret Zoph, and Noam Shazeer. 2021.
\newblock Switch transformers: Scaling to trillion parameter models with simple
  and efficient sparsity.
\newblock \emph{arXiv preprint arXiv:2101.03961}.

\bibitem[{Ge et~al.(2013)Ge, He, Ke, and Sun}]{ge2013optimized}
Tiezheng Ge, Kaiming He, Qifa Ke, and Jian Sun. 2013.
\newblock Optimized product quantization for approximate nearest neighbor
  search.
\newblock In \emph{Proceedings of the IEEE Conference on Computer Vision and
  Pattern Recognition}, pages 2946--2953.

\bibitem[{Graves et~al.(2014)Graves, Wayne, and Danihelka}]{graves2014neural}
Alex Graves, Greg Wayne, and Ivo Danihelka. 2014.
\newblock Neural turing machines.
\newblock \emph{arXiv preprint arXiv:1410.5401}.

\bibitem[{Guu et~al.(2020)Guu, Lee, Tung, Pasupat, and Chang}]{guu2020realm}
Kelvin Guu, Kenton Lee, Zora Tung, Panupong Pasupat, and Ming-Wei Chang. 2020.
\newblock Realm: Retrieval-augmented language model pre-training.
\newblock \emph{arXiv preprint arXiv:2002.08909}.

\bibitem[{Hendrycks and Gimpel(2016)}]{hendrycks2016gaussian}
Dan Hendrycks and Kevin Gimpel. 2016.
\newblock Gaussian error linear units (gelus).
\newblock \emph{arXiv preprint arXiv:1606.08415}.

\bibitem[{Huang et~al.(2021)Huang, Sainath, Peyser, Kumar, Rybach, and
  Strohman}]{huang2021lookup}
W~Ronny Huang, Tara~N Sainath, Cal Peyser, Shankar Kumar, David Rybach, and
  Trevor Strohman. 2021.
\newblock Lookup-table recurrent language models for long tail speech
  recognition.
\newblock \emph{arXiv preprint arXiv:2104.04552}.

\bibitem[{Jegou et~al.(2011)Jegou, Douze, and Schmid}]{jegou2011product}
Herve Jegou, Matthijs Douze, and Cordelia Schmid. 2011.
\newblock Product quantization for nearest neighbor search.
\newblock \emph{IEEE transactions on pattern analysis and machine
  intelligence}, 33(1):117--128.

\bibitem[{Kaiser et~al.(2018)Kaiser, Roy, Vaswani, Pamar, Bengio, Uszkoreit,
  and Shazeer}]{kaiser2018fast}
{\L}ukasz Kaiser, Aurko Roy, Ashish Vaswani, Niki Pamar, Samy Bengio, Jakob
  Uszkoreit, and Noam Shazeer. 2018.
\newblock Fast decoding in sequence models using discrete latent variables.
\newblock \emph{arXiv preprint arXiv:1803.03382}.

\bibitem[{Katz(1987)}]{katz1987estimation}
Slava Katz. 1987.
\newblock Estimation of probabilities from sparse data for the language model
  component of a speech recognizer.
\newblock \emph{IEEE transactions on acoustics, speech, and signal processing},
  35(3):400--401.

\bibitem[{Khandelwal et~al.(2019)Khandelwal, Levy, Jurafsky, Zettlemoyer, and
  Lewis}]{khandelwal2019generalization}
Urvashi Khandelwal, Omer Levy, Dan Jurafsky, Luke Zettlemoyer, and Mike Lewis.
  2019.
\newblock Generalization through memorization: Nearest neighbor language
  models.
\newblock \emph{arXiv preprint arXiv:1911.00172}.

\bibitem[{Kingma and Ba(2015)}]{adam}
Diederik~P. Kingma and Jimmy Ba. 2015.
\newblock \href {http://arxiv.org/abs/1412.6980} {Adam: {A} method for
  stochastic optimization}.
\newblock In \emph{3rd International Conference on Learning Representations,
  {ICLR} 2015, San Diego, CA, USA, May 7-9, 2015, Conference Track
  Proceedings}.

\bibitem[{Kneser and Ney(1995)}]{kneser1995improved}
Reinhard Kneser and Hermann Ney. 1995.
\newblock Improved backing-off for m-gram language modeling.
\newblock In \emph{1995 international conference on acoustics, speech, and
  signal processing}, volume~1, pages 181--184. IEEE.

\bibitem[{Koehn(2009)}]{koehn2009statistical}
Philipp Koehn. 2009.
\newblock \emph{Statistical machine translation}.
\newblock Cambridge University Press.

\bibitem[{Krishna et~al.(2021)Krishna, Roy, and Iyyer}]{krishna2021hurdles}
Kalpesh Krishna, Aurko Roy, and Mohit Iyyer. 2021.
\newblock Hurdles to progress in long-form question answering.
\newblock \emph{arXiv preprint arXiv:2103.06332}.

\bibitem[{Lample et~al.(2019)Lample, Sablayrolles, Ranzato, Denoyer, and
  J{\'e}gou}]{lample2019large}
Guillaume Lample, Alexandre Sablayrolles, Marc'Aurelio Ranzato, Ludovic
  Denoyer, and Herv{\'e} J{\'e}gou. 2019.
\newblock Large memory layers with product keys.
\newblock \emph{arXiv preprint arXiv:1907.05242}.

\bibitem[{Lepikhin et~al.(2020)Lepikhin, Lee, Xu, Chen, Firat, Huang, Krikun,
  Shazeer, and Chen}]{lepikhin2020gshard}
Dmitry Lepikhin, HyoukJoong Lee, Yuanzhong Xu, Dehao Chen, Orhan Firat, Yanping
  Huang, Maxim Krikun, Noam Shazeer, and Zhifeng Chen. 2020.
\newblock Gshard: Scaling giant models with conditional computation and
  automatic sharding.
\newblock \emph{arXiv preprint arXiv:2006.16668}.

\bibitem[{Lewis et~al.(2020)Lewis, Perez, Piktus, Petroni, Karpukhin, Goyal,
  K{\"u}ttler, Lewis, Yih, Rockt{\"a}schel et~al.}]{lewis2020retrieval}
Patrick Lewis, Ethan Perez, Aleksandra Piktus, Fabio Petroni, Vladimir
  Karpukhin, Naman Goyal, Heinrich K{\"u}ttler, Mike Lewis, Wen-tau Yih, Tim
  Rockt{\"a}schel, et~al. 2020.
\newblock Retrieval-augmented generation for knowledge-intensive nlp tasks.
\newblock \emph{arXiv preprint arXiv:2005.11401}.

\bibitem[{Mikolov et~al.(2013)Mikolov, Chen, Corrado, and
  Dean}]{mikolov2013efficient}
Tomas Mikolov, Kai Chen, Greg Corrado, and Jeffrey Dean. 2013.
\newblock Efficient estimation of word representations in vector space.
\newblock \emph{arXiv preprint arXiv:1301.3781}.

\bibitem[{Panigrahy et~al.(2021)Panigrahy, Wang, and
  Zaheer}]{panigrahy2021sketch}
Rina Panigrahy, Xin Wang, and Manzil Zaheer. 2021.
\newblock Sketch based memory for neural networks.
\newblock In \emph{International Conference on Artificial Intelligence and
  Statistics}, pages 3169--3177. PMLR.

\bibitem[{Radford et~al.(2018)Radford, Narasimhan, Salimans, and
  Sutskever}]{radford2018improving}
Alec Radford, Karthik Narasimhan, Tim Salimans, and Ilya Sutskever. 2018.
\newblock Improving language understanding by generative pre-training.
\newblock \emph{URL https://s3-us-west-2. amazonaws.
  com/openai-assets/research-covers/languageunsupervised/language understanding
  paper. pdf}.

\bibitem[{Raffel et~al.(2019)Raffel, Shazeer, Roberts, Lee, Narang, Matena,
  Zhou, Li, and Liu}]{raffel2019exploring}
Colin Raffel, Noam Shazeer, Adam Roberts, Katherine Lee, Sharan Narang, Michael
  Matena, Yanqi Zhou, Wei Li, and Peter~J Liu. 2019.
\newblock Exploring the limits of transfer learning with a unified text-to-text
  transformer.
\newblock \emph{arXiv preprint arXiv:1910.10683}.

\bibitem[{Roy et~al.(2021)Roy, Saffar, Vaswani, and
  Grangier}]{roy2021efficient}
Aurko Roy, Mohammad Saffar, Ashish Vaswani, and David Grangier. 2021.
\newblock Efficient content-based sparse attention with routing transformers.
\newblock \emph{Transactions of the Association for Computational Linguistics},
  9:53--68.

\bibitem[{Roy et~al.(2018)Roy, Vaswani, Neelakantan, and
  Parmar}]{roy2018theory}
Aurko Roy, Ashish Vaswani, Arvind Neelakantan, and Niki Parmar. 2018.
\newblock Theory and experiments on vector quantized autoencoders.
\newblock \emph{arXiv preprint arXiv:1805.11063}.

\bibitem[{Schneider et~al.(2019)Schneider, Baevski, Collobert, and
  Auli}]{schneider2019wav2vec}
Steffen Schneider, Alexei Baevski, Ronan Collobert, and Michael Auli. 2019.
\newblock wav2vec: Unsupervised pre-training for speech recognition.
\newblock \emph{arXiv preprint arXiv:1904.05862}.

\bibitem[{Shazeer et~al.(2017)Shazeer, Mirhoseini, Maziarz, Davis, Le, Hinton,
  and Dean}]{shazeer2017outrageously}
Noam Shazeer, Azalia Mirhoseini, Krzysztof Maziarz, Andy Davis, Quoc Le,
  Geoffrey Hinton, and Jeff Dean. 2017.
\newblock Outrageously large neural networks: The sparsely-gated
  mixture-of-experts layer.
\newblock \emph{arXiv preprint arXiv:1701.06538}.

\bibitem[{So et~al.(2021)So, Ma{\'n}ke, Liu, Dai, Shazeer, and
  Le}]{so2021primer}
David~R So, Wojciech Ma{\'n}ke, Hanxiao Liu, Zihang Dai, Noam Shazeer, and
  Quoc~V Le. 2021.
\newblock Primer: Searching for efficient transformers for language modeling.
\newblock \emph{arXiv preprint arXiv:2109.08668}.

\bibitem[{Su et~al.(2021)Su, Lu, Pan, Wen, and Liu}]{su2021roformer}
Jianlin Su, Yu~Lu, Shengfeng Pan, Bo~Wen, and Yunfeng Liu. 2021.
\newblock Roformer: Enhanced transformer with rotary position embedding.
\newblock \emph{arXiv preprint arXiv:2104.09864}.

\bibitem[{Sun and Iyyer(2021)}]{sun2021revisiting}
Simeng Sun and Mohit Iyyer. 2021.
\newblock Revisiting simple neural probabilistic language models.
\newblock \emph{arXiv preprint arXiv:2104.03474}.

\bibitem[{Thorup(2015)}]{thorup2015high}
Mikkel Thorup. 2015.
\newblock High speed hashing for integers and strings.
\newblock \emph{arXiv preprint arXiv:1504.06804}.

\bibitem[{van~den Oord et~al.(2017)van~den Oord, Vinyals, and
  Kavukcuoglu}]{vqvae}
A{\"{a}}ron van~den Oord, Oriol Vinyals, and Koray Kavukcuoglu. 2017.
\newblock \href {http://arxiv.org/abs/1711.00937} {Neural discrete
  representation learning}.
\newblock \emph{CoRR}, abs/1711.00937.

\bibitem[{Vaswani et~al.(2017)Vaswani, Shazeer, Parmar, Uszkoreit, Jones,
  Gomez, Kaiser, and Polosukhin}]{vaswani2017attention}
Ashish Vaswani, Noam Shazeer, Niki Parmar, Jakob Uszkoreit, Llion Jones,
  Aidan~N Gomez, {\L}ukasz Kaiser, and Illia Polosukhin. 2017.
\newblock Attention is all you need.
\newblock In \emph{Advances in neural information processing systems}, pages
  5998--6008.

\bibitem[{Wang et~al.(2019)Wang, Pruksachatkun, Nangia, Singh, Michael, Hill,
  Levy, and Bowman}]{wang2019superglue}
Alex Wang, Yada Pruksachatkun, Nikita Nangia, Amanpreet Singh, Julian Michael,
  Felix Hill, Omer Levy, and Samuel Bowman. 2019.
\newblock Superglue: A stickier benchmark for general-purpose language
  understanding systems.
\newblock \emph{Advances in neural information processing systems}, 32.

\bibitem[{Wang et~al.(2018)Wang, Stanton, Zhang, Ryan, Battenberg, Shor, Xiao,
  Jia, Ren, and Saurous}]{wang2018style}
Yuxuan Wang, Daisy Stanton, Yu~Zhang, RJ-Skerry Ryan, Eric Battenberg, Joel
  Shor, Ying Xiao, Ye~Jia, Fei Ren, and Rif~A Saurous. 2018.
\newblock Style tokens: Unsupervised style modeling, control and transfer in
  end-to-end speech synthesis.
\newblock In \emph{International Conference on Machine Learning}, pages
  5180--5189. PMLR.

\bibitem[{Weston et~al.(2014)Weston, Chopra, and Bordes}]{weston2014memory}
Jason Weston, Sumit Chopra, and Antoine Bordes. 2014.
\newblock Memory networks.
\newblock \emph{arXiv preprint arXiv:1410.3916}.

\bibitem[{Wu et~al.(2021)Wu, Xiao, Codella, Liu, Dai, Yuan, and
  Zhang}]{wu2021cvt}
Haiping Wu, Bin Xiao, Noel Codella, Mengchen Liu, Xiyang Dai, Lu~Yuan, and Lei
  Zhang. 2021.
\newblock Cvt: Introducing convolutions to vision transformers.
\newblock \emph{arXiv preprint arXiv:2103.15808}.

\bibitem[{Wu et~al.(2016)Wu, Schuster, Chen, Le, Norouzi, Macherey, Krikun,
  Cao, Gao, Macherey et~al.}]{wu2016google}
Yonghui Wu, Mike Schuster, Zhifeng Chen, Quoc~V Le, Mohammad Norouzi, Wolfgang
  Macherey, Maxim Krikun, Yuan Cao, Qin Gao, Klaus Macherey, et~al. 2016.
\newblock Google's neural machine translation system: Bridging the gap between
  human and machine translation.
\newblock \emph{arXiv preprint arXiv:1609.08144}.

\bibitem[{Yu et~al.(2018)Yu, Yuan, Fang, and Jin}]{yu2018product}
Tan Yu, Junsong Yuan, Chen Fang, and Hailin Jin. 2018.
\newblock Product quantization network for fast image retrieval.
\newblock In \emph{Proceedings of the European Conference on Computer Vision
  (ECCV)}, pages 186--201.

\bibitem[{Zaheer et~al.(2020)Zaheer, Guruganesh, Dubey, Ainslie, Alberti,
  Ontanon, Pham, Ravula, Wang, Yang et~al.}]{zaheer2020big}
Manzil Zaheer, Guru Guruganesh, Kumar~Avinava Dubey, Joshua Ainslie, Chris
  Alberti, Santiago Ontanon, Philip Pham, Anirudh Ravula, Qifan Wang, Li~Yang,
  et~al. 2020.
\newblock Big bird: Transformers for longer sequences.
\newblock In \emph{NeurIPS}.

\end{thebibliography}
\bibliographystyle{acl_natbib}


\end{document}